\documentclass{article}
\usepackage{spconf,amsmath,graphicx}
\usepackage[hyphens]{url}
\usepackage{breakurl}
\usepackage{hyperref}
\usepackage{pgfplots}
\usepackage{mathtools}
\usepackage{booktabs}
\pgfplotsset{compat = newest} 

\renewenvironment{enumerate}%
  {\begin{list}{\arabic{enumi}.}%
     {\topsep=0in\itemsep=0in\parsep=0in\partopsep=0in\usecounter{enumi}}%
   }{\end{list}}

\usetikzlibrary{decorations.markings,calc}

\title{\vspace{-1.4cm}Scaling Recurrent Neural Network Language Models}
\name{Will Williams, Niranjani Prasad, David Mrva, Tom Ash, Tony Robinson   \thanks{This paper summarises the results of SMART award "Large scale neural network language models" which ran from March 2013 to August 2014.}}
\address{Cantab Research, Cambridge, UK\\
\tt{\{willw,tonyr\}@cantabResearch.com}
}

\begin{document}
%
\maketitle
\begin{abstract}
This paper investigates the scaling properties of Recurrent Neural Network Language Models (RNNLMs). We discuss how to train very large RNNs on GPUs and address the questions of how RNNLMs scale with respect to model size, training-set size, computational costs and memory. Our analysis shows that despite being more costly to train, RNNLMs obtain much lower perplexities on standard benchmarks than $n$-gram models. We train the largest known RNNs and present relative word error rates gains of 18\% on an ASR task. We also present the new lowest perplexities on the recently released billion word language modelling benchmark, 1 BLEU point gain on machine translation and a 17\% relative hit rate gain in word prediction.
\end{abstract}
\section{Introduction}
\vspace{-0.15cm}
\label{sec:intro}
Statistical language models form a crucial component of many applications such as automatic speech recognition (ASR), machine translation (MT) and prediction for mobile phone text input. One such class of models, Recurrent Neural Network Language Models (RNNLMs), provide a rich and powerful way to model sequential language data. Despite an initial flurry of interest in RNNs in the early '90s for acoustic modelling \cite{robinson, williams, schuster}, the computational cost and memory overheads of training large RNNs proved prohibitive. Since the earliest days of large vocabulary speech recognition, \emph{n}-gram language models have been the dominant paradigm. However, recent work by Mikolov \cite{mikolov} on RNNLMs has shown that modestly sized RNNLMs can now be trained and have been shown to be competitive with \emph{n}-grams. Mikolov trained RNNLMs with 800 hidden state units; Google's language modelling benchmark \cite{goog1bn} subsequently established baseline RNNLM results with 1024 hidden state units. Many of the most recent ASR evaluations involve RNNLMs, highlighting their widespread popularity.

Although RNNs have replaced \emph{n}-grams as state-of-the-art, the question remains whether these architectures will scale. Single CPU-based implementations have suffered from computational limitations and have been unable to scale to the large number of parameters now commonplace in the neural net literature. Concretely, we address the questions of how RNNLMs scale with respect to model size, training-set size, processing power and memory. Our primary performance metric here is perplexity and therefore all discussion on performance will relate specifically to the ability to reduce perplexity. The results presented here show the largest reductions in perplexity reported so far over KN 5-grams.
\vspace{-0.25cm}
\section{Data sets}
\vspace{-0.15cm}
\subsection{Training Data}
The main training corpus we use throughout this paper is an internally collated and processed data set. All data predates Nov. 2013 and totals approximately \textbf{8bn words}, comprising:
\begin{enumerate}
\item \textbf{[880m] \emph{Spoken News}}: Derived from transcripts of radio and news shows.
\item \textbf{[1.7bn] \emph{Wikipedia}}: Copy of Wikipedia articles.
\item \textbf{[6.1bn] \emph{Written News}}: Derived from a web crawl of many popular news sites.
\end{enumerate}

We used a rolling buffer of minimum size of 140 characters split on sentences to deduplicate the corpora. We used multiple hash functions to store past occurrences efficiently and excluded a negligibly small proportion of text due to false positives. We then put the corpus through a typical ASR normalisation and tokenisation pipeline - expanding out numbers into digit strings, removing most punctuation, casting to lower case and inserting sentence markers.
\vspace{-0.15cm}
\subsection{Test Data}
Unless otherwise stated, all our testing was done on a standard test benchmark: TED test data IWSLT14.SLT.tst2010\footnote{\url{http://hltshare.fbk.eu/IWSLT2014/IWSLT14.SLT.tst2010.en-fr.en.tgz}}. We chose this out-of-domain test set to exclude any possibility of text overlap between training and test sets, and also because of the availability of a publicly available KALDI recipe\footnote{Kaldi Revision 4084, recipe: \url{http://sourceforge.net/p/kaldi/code/HEAD/tree/trunk/egs/tedlium/s5/}}.
\vspace{-0.15cm}
\subsection{Entropy Filtering}
To filter our corpus we used cross-entropy difference scoring \cite{klakow,moorelewis}. Typically, sentences are used as the natural choice for segments in this style of filtering.  However, for the composite corpus, we wanted to maintain between-sentence context for the benefit of the RNNLM training, and so did not want to filter our corpus on a sentence by sentence basis. We instead implemented a rolling-buffer solution: a cross-entropy difference score was calculated across rolling batches of 16 sentences.  If a sentence was ever part of a rolling buffer with cross-entropy difference below the threshold, it was maintained in the output corpus.  This led to output that was more likely to be drawn from consecutive sentences, maintaining between-sentence information.  The entropy threshold was then set at empirically determined levels to produce filtered corpora of the desired size, i.e. $\frac{1}{2}$, $\frac{1}{4}$ etc. of the original corpus size, as determined by word count.  For typical in-domain data we used IWSLT13.ASR.train.en\footnote{\url{http://hltshare.fbk.eu/IWSLT2013/IWSLT13.ASR.train.en}}.
\vspace{-0.25cm}
\section{Scaling Properties of N-grams}
\vspace{-0.15cm}
\emph{n}-gram language models have maintained their dominance in statistical language modelling largely due to their simplicity and ease of computation. Although \emph{n}-grams can be modelled and queried comparatively fast, they exhibit diminishing gains when built on more data.

\begin{figure}[h!]
\centering
\begin{tikzpicture}[scale=0.8]
\begin{axis}[
xlabel = Training Words,
ylabel = Perplexity,
xmode=log,
ymode=log,
ymajorgrids=true,
      yticklabel={
        \pgfkeys{/pgf/fpu=true}
        \pgfmathparse{exp(\tick)}%
        \pgfmathprintnumber[fixed, precision=0]{\pgfmathresult}
        \pgfkeys{/pgf/fpu=false}
      },
ytick={70,100,125,150,175},
grid style=dashed
]

\addplot coordinates {
( 125000000, 170.845 )
( 250000000, 157.477 )
( 500000000, 147.658 )
( 1000000000, 137.378 )
( 2000000000, 130.662 )
( 4000000000, 123.083 )
( 8000000000, 117.621 )
};
\addplot[dashed] coordinates{
( 8000000000, 117.621 )
( 16000000000, 113.13 )
( 32000000000, 108.711 )
( 64000000000, 104.851 )
( 128000000000, 101.465 )
( 256000000000, 98.488 )
( 512000000000, 95.861 )
( 1024000000000, 93.456 )
( 2048000000000, 91.389 )
( 4096000000000, 89.553 )
( 10000000000000, 87.481 )
};

\addlegendentry{5-gram-RS};
\draw (axis cs:0,1) -- ({axis cs:0,1}-|{rel axis cs:1,0});

\end{axis}
\end{tikzpicture}
\vspace{-0.25cm}
\caption{Effect of increasing training data size on a randomly selected (RS) subsets of our training corpus with a \emph{5}-gram. The dashed line is extrapolation.}
\label{ngramscale}
\end{figure}
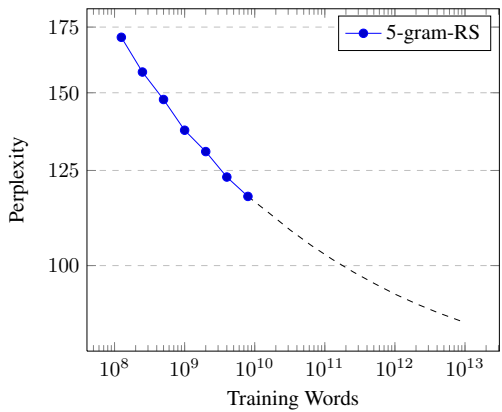

\emph{n}-grams can be computed on larger data sets more easily than RNNLMs but as the extrapolation in Figure~\ref{ngramscale} indicates, each order of magnitude increase in the training data above ${10}^{12}$ words gives a reduction in perplexity of less than 6\%. The largest current text corpora such as Google \emph{n}-grams\footnote{\url{http://googleresearch.blogspot.co.uk/2006/08/all-our-n-gram-are-belong-to-you.html}} and CommonCrawl \cite{commoncrawl} are about size ${10}^{12}$.  The asymptote of an exponential curve fitted to  Figure~\ref{ngramscale} is approximately perplexity 73; these values represent a hard limit on the performance of 5-grams on this test set. \emph{n}-grams also scale very poorly with respect to memory. At 8bn words the KN 5-gram already takes up 362GB in ARPA format and 69GB in KenLM \cite{kenlm} binary trie format - already impractically large for current commercial ASR. In comparison to RNNLMs, \emph{n}-grams take up massively more space for modest entropy improvements and therefore do not scale well with respect to data size.

Additionally, increasing the order of the \emph{n}-gram model gives vanishingly small gains for anything above a 5-gram trained on currently available corpora. The number of parameters of an \emph{n}-gram increases significantly with its order, making attempts to increase the order impractical.
\vspace{-0.25cm}
\section{RNNLMs on GPUs}
\vspace{-0.15cm}
\subsection{Background}
Recent attempts to train very large networks (on the order of billions of parameters) have required many CPU cores to work in concert \cite{goog1bn,distbelief}. High-end Graphics Processing Units (GPUs) represent a viable alternative, being both affordable and capable of extremely high computational throughput. GPUs have therefore been one of the key elements in the resurgence of neural networks over the last decade. RNNLMs are highly amenable to parallelisation on GPUs due to their predominant reliance on linear algebra operations (such as matrix multiplies) for which highly optimised GPU-based libraries are publicly available.

In contrast to \emph{n}-gram models, which simply count the occurrences of particular combinations of words, RNNs learn a distributed hidden representation for each context from which a probability distribution over all words in a given vocabulary can be obtained.




We use a standard RNN architecture \cite{mikolovStandard} but dispense with bias units to maximise efficiency on the GPU and because, in our experience, they do not provide any practical benefit. Recent work has improved our understanding of how to effectively train RNNs \cite{bengioAdvances}. However, many of these improvements - such as optimisation and regularisation techniques - make large claims on resources. Very large RNNs with a large number of hidden state units, \emph{nstate}, can only be trained efficiently on current generation GPU hardware if one simplifies both the architecture and training algorithm as far as possible. We believe our setup from 2013 in this paper gives better speedups than previously reported elsewhere \cite{mikolov2011rnnlm,sutskever2011generating,huang2013accelerating,jeff,li2014large}.
\vspace{-0.15cm}
\subsection{Implementation}
To achieve high throughput and utilisation on GPUs we train a standard RNN with stochastic gradient descent and rmsprop\cite{rmsprop}.
Where possible we use CUDA's highly optimised linear algebra library cuBLAS to make SGEMM calls. Where this wasn't possible we wrote and optimised our own custom CUDA kernels.
We use floats everywhere; the precision which doubles offer is unnecessary to learn good representations.
We pack our input data using a data offset scheme which indexes the input corpus at a number of different points, denoted \emph{noffset}. This entails managing \emph{noffset} by minibatch size different hidden states throughout training. Typically we use \emph{noffset} 128, similar to \cite{jeff}.
Empirically, we found a minibatch size of {\raise.17ex\hbox{$\scriptstyle\mathtt{\sim}$}} 8-10 represents a good trade-off between memory usage and perplexity. It also allows us to marshal a large and very efficient matrix multiply to perform all the hidden state-to-state and gradient calculations in one cuBLAS SGEMM operation. 
We use a very large rmsprop smoothing constant (0.9995). For the input-to-state and state-to-output matrices we approximate the majority of the rmsprop values by averaging over \emph{nstate} units such that each word has just one rmsprop value rather than $n$ values. This almost halves the total memory usage.
We train using Noise Contrastive Estimation~\cite{nce}.   When reporting perplexities we explicitly normalise the output distribution.  When performing lattice rescoring we find empirically that the small amount of normalisation noise does not significantly change the accuracy, whilst considerably faster than standard class based modelling.

\begin{figure}
\begin{tikzpicture}
\vspace{-0.25cm}
\begin{axis}[
xlabel = Training words (bn),
ylabel = Perplexity,
ymode=log,
xmode=log,
xtick=data,
ytick={40,60,80, 100, 120,140,160},
yticklabel={
\pgfkeys{/pgf/fpu=true}
\pgfmathparse{exp(\tick)}%
\pgfmathprintnumber[fixed, precision=0]{\pgfmathresult}
\pgfkeys{/pgf/fpu=false}
},
xticklabel={
\pgfkeys{/pgf/fpu=true}
\pgfmathparse{exp(\tick)}%
\pgfmathprintnumber[fixed, precision=2]{\pgfmathresult}
\pgfkeys{/pgf/fpu=false}
},
legend image post style={xscale=0.5},
legend style={at={(0.6 ,0.55)},anchor=north west},
ymajorgrids=true,
grid style=dashed
]

\addplot coordinates {
( 0.25, 157.477 )
( 0.5, 147.658 )
( 1, 137.378 )
( 2, 130.662 )
( 4, 123.083 )
( 8, 117.621 )
};
\addlegendentry{5-gram-RS} ;

\addplot coordinates {
( 0.25, 120.017 )
( 0.5, 111.985 )
( 1, 107.855 )
( 2, 106.459 )
( 4, 109.131 )
( 8, 117.994 )
};
\addlegendentry{5-gram-EF} ;

\addplot[mark=square] coordinates {
( 0.25, 83.6101 )
( 0.5, 75.4577 )
( 1, 72.5293 )
( 2, 69.324 )
( 4, 65.7581 )
( 8, 66.2034 )
};
\addlegendentry{RNN-2048-RS} ;

\addplot coordinates {
( 0.25, 68.1386)
( 0.5, 65.5026 )
( 1, 61.7875 )
( 2, 60.369 )
( 4, 60.7983 )
( 8, 66.2034 )
};
\addlegendentry{RNN-2048-EF} ;

\end{axis}
\end{tikzpicture}
\vspace{-0.3cm}
\caption{Performance of \emph{5}-grams against \emph{nstate} 2048 RNNs with increasing training data size. We test on Randomly Selected (RS) splits and Entropy Filtered (EF) splits of the 8bn corpus.}
\label{spandata}
\end{figure}
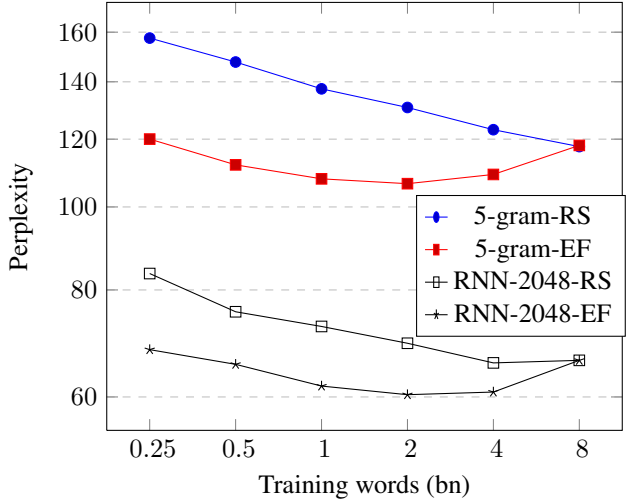

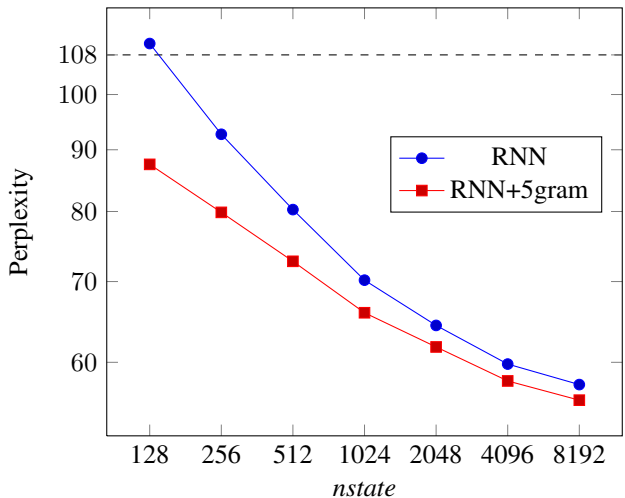
\begin{figure}
\begin{tikzpicture}
\begin{axis}[
xlabel = \emph{nstate},
ylabel = Perplexity,
ymode=log,
xmode=log,
xtick=data,
ytick={50,60,70,80,90,100},
xticklabels={128,256,512,1024,2048,4096,8192},
yticklabel={
\pgfkeys{/pgf/fpu=true}
\pgfmathparse{exp(\tick)}%
\pgfmathprintnumber[fixed, precision=0]{\pgfmathresult}
\pgfkeys{/pgf/fpu=false}
},
grid style={dashed,black},
extra y ticks={107.855},
extra tick style={grid=major},
legend style={at={(0.55 ,0.7)},anchor=north west}
]

\addplot coordinates {
( 128, 110.215 )
( 256, 92.7 )
( 512, 80.2907 )
( 1024, 70.1848 )
( 2048, 64.377 ) 
( 4096, 59.8006 )
( 8192, 57.5028 )
};
\addlegendentry{RNN} ;

\addplot coordinates {
( 128, 87.5275 )
( 256, 79.8797 )
( 512, 72.7577 )
( 1024, 65.9488 )
( 2048, 61.7875 )
( 4096, 57.9057 )
( 8192, 55.8196 )
};
\addlegendentry{RNN+5gram} ;

\end{axis}
\end{tikzpicture}
\vspace{-0.25cm}
\caption{Scaling \emph{nstate} trained on 1bn words of the entropy filtered 8bn corpus. Dashed line is the 5-gram baseline.}
\label{spanstates}
\vspace{-0.4cm}
\end{figure}
\vspace{-0.2cm}
\subsection{Analysis}
Figure~\ref{spanstates} shows that as we scale \emph{nstate} we observe a near linear reduction in log perplexity with log training words. Given that \emph{n}-grams scale very poorly with respect to their order it is clear that, on a fixed-size data set, RNNs scale much better with model size.
For network sizes over \emph{nstate} 1024, our implementations on an Nvidia GeForce GTX Titan give 100x speedups against the baseline single core Mikolov implementation\footnote{\url{http://rnnlm.org/}} on a 3.4GHz Intel i7 CPU. Despite much improved training times on the GPU, our larger RNNs take on the order of days to train rather than hours which \emph{n}-grams require. However, more compute power will favour RNNs since larger \emph{nstate} RNNs can leverage the extra computation to scale the model size with \emph{nstate} - something which yields a much smaller performance gain for \emph{n}-grams.
With respect to scaling the training set size, Figure~\ref{spandata} shows that the \emph{nstate} 2048 saturates (i.e. can not make good use of more data) on 4bn words when trained on a randomly chosen splits. We can, however, increase the model size (i.e. \emph{nstate}) to mitigate this problem - an approach which, as already discussed, is impractical for \emph{n}-grams.
With an \emph{nstate} 8192 RNN from Figure~\ref{spanstates} we have already reduced perplexity to 57.5, which is below the 73 we believe to be the asymptotic minimum for a 5-gram on this task.  It therefore seems unlikely that even a 5-gram with unlimited data and a further 15\% reduction from entropy filtering could ever outperform an RNNLM.
In addition, the \emph{nstate} 8192 RNN corresponds to a 48\% reduction over the 5-gram. Despite being so much better in terms of perplexity on the same amount of data, the RNN uses only 886M parameters - approximately half the number of the 5-gram. Moreover, we can match the perplexity of the 5-gram with an \emph{nstate} 128 RNN which uses under 1\% of the parameters. In summary, although training time is much larger for an RNN and we have to increase \emph{nstate} when scaling the training set size, RNNs make much better use of the additional data than \emph{n}-grams and use far fewer parameters to do so. 
\vspace{-0.25cm}
\section{Results}
\vspace{-0.15cm}
\subsection{Language Modelling}
\vspace{-0.25cm}
\begin{table}
\begin{tabular}{@{}cccc@{}}
\toprule
\multicolumn{1}{l}{\textbf{Model}} & \textbf{\begin{tabular}[c]{@{}c@{}}\# Params \\ {[}millions{]}\end{tabular}} & \textbf{Training Time} & \textbf{Perplexity}    \\ \midrule
KN 5-gram                        & 1,740                                                                           & 30m                   & 66.9                     \\
RNN - 128                        & 16.4                                                                           & 6h                   & 60.8                     \\
RNN - 256                        & 32.8                                                                           & 16h                   & 57.3                     \\
RNN - 512                        & 65.8                                                                           & 1d2h                   & 53.2                     \\
RNN - 1024                        & 132                                                                           & 2d2h                   & 48.9                     \\
RNN - 2048                        & 266                                                                           & 4d7h                   & 45.2                     \\
RNN - 4096                        & 541                                                                           & 14d5h                   & 42.4                     \\ \bottomrule
\end{tabular}
\caption{Perplexities on shard-0 ('news.en.heldout-00000-of-00050') from \cite{goog1bn}.}
\label{google1bntable}
\vspace{-0.25cm}
\end{table}

We present our RNNLM results on the recently released billion word language modelling benchmark\cite{goog1bn}. The vocabulary size for this task is very large (770K) - we therefore train the RNNs on a much smaller 64k vocabulary and interpolate them with a full size 5-gram to fill in rare word probabilities. RNN perplexities in Table \ref{google1bntable} are interpolated with the KN 5-gram. The \emph{nstate} 4096 RNN is larger in state size than those in \cite{goog1bn} and we achieve a better perplexity (42.4) with that one single model than the combination of interpolated models in \cite{goog1bn} whilst also using only 3\% of the total parameters. 
\vspace{-0.3cm}
\subsection{ASR}
\begin{figure}
\centering
\begin{tikzpicture}
\begin{axis}[
xlabel = \emph{nstate},
ylabel = WER \%,
symbolic x coords={n/a,128,256,512,1024,2048,4096,8192},
xtick={n/a,128,256,512,1024,2048,4096,8192},
xticklabel style={text height=1.5ex},
ymax=16,
legend style={at={(1,1)}} 
]

\addplot[mark=square] coordinates {
(n/a, 13.75 )
(128, 12.81 )
(256, 12.55 )
(512, 12.36 )
(1024, 12.19 )
(2048, 12.34 )
};
\addlegendentry{IWSLT-unseg};
\addplot coordinates {
(n/a, 12.8 )
(128, 12.0 )
(256, 11.7 )
(512, 11.5 )
(1024, 11.5 )
(2048, 11.5 )
};
\addlegendentry{IWSLT-seg};
\addplot coordinates {
(n/a, 11.16 )
(128, 10.47 )
(256, 10.12 )
(512, 10.01 )
(1024, 9.81 )
(2048, 9.83 )
(4096, 9.42 )
(8192, 9.3 )
};
\addlegendentry{en-US-unseg};

\addplot coordinates {
(n/a, 10.3 )
(128, 9.8 )
(256, 9.5 )
(512, 9.2 )
(1024, 9.1 )
(2048, 8.8 )
(4096, 8.6 )
(8192, 8.5 )
};
\addlegendentry{en-US-seg};

\addplot coordinates {
(n/a, 10.15 )
(128, 9.26 )
(256, 8.96 )
(512, 9.02 )
(1024, 8.62 )
(2048, 8.62 )
(4096, 8.45 )
(8192, 8.32 )
};
\addlegendentry{SM-unseg};

\end{axis}
\end{tikzpicture}
\vspace{-0.25cm}
\caption{Rescoring Kaldi Lattices with RNNLMs. `n/a' refers to rescoring with \emph{n}-gram only and is the \emph{n}-gram baseline.}
\label{asr}
\vspace{-0.25cm}
\end{figure}
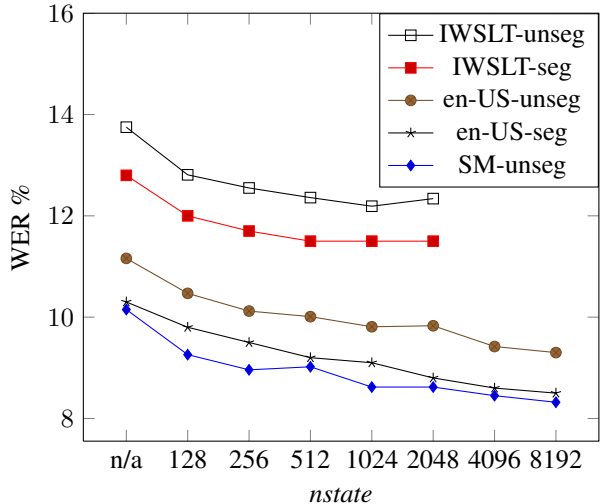
We evaluated the RNNLMs by rescoring lattices on three different systems, using the IWSLT14.SLT.tst2010 data both with and without the supplied segmentation. The `IWSLT' system uses the Kaldi TEDLIUM~\cite{tedlium} recipe for acoustic models and language models built on~\cite{goog1bn}\footnote{The 12.8\% IWSLT result is a competitive baseline that complies with IWSLT rules, using freely redistributable sources and can be recreated from the Kaldi TEDLIUM s5 recipe and \url{http://cantabResearch.com/cantab-TEDLIUM.tar.bz2}.}. The \emph{nstate} 2048 models overtrain on the small entropy filtered corpus. The `en-US' system uses the same framework but with the RNNLMs from Figure~\ref{spanstates} and internal acoustic models. The `SM' system also uses the RNNLMs from Figure~\ref{spanstates}, but within the commercial service available at \url{speechmatics.com}. Lattices were rescored using a highly efficient internal lattice rescoring tool which operates in considerably less than real time.  Over both the `en-US' and `SM' task we see an average reduction in WER of 18\% relative to rescoring with the \emph{n}-gram alone by using \emph{nstate} 8192 RNNLMs.
\vspace{-0.3cm}
\subsection{Machine Translation}
In our MT experiments we rescored the WFST lattices from the WMT13 EN-RU system developed at Cambridge University Engineering Department \cite{juan}. The evaluation measure commonly used in statistical machine translation, BLEU score, was 32.34 with our baseline \emph{n}-gram model and increased to 33.45 with an \emph{nstate} 3520 RNNLM (bigger is better for BLEU scores). The improvement of 1 BLEU point over the baseline demonstrates the utility of RNNLMs in not just ASR but also statistical MT systems.
\vspace{-0.3cm}
\subsection{Word prediction}
Word prediction involves predicting the next word in a string, with performance determined by percentage of times the target word is in the top 3 words predicted (known as `hit-rate'). For this task we train on 100M words from the BNC corpus~\cite{bnc} using a vocabulary size of 10K and the RNN using a shortlist of 100 candidate words generated from a heavily pruned 2~MB \emph{n}-gram.
For our task of word prediction on mobile phones we have tight memory restrictions and therefore choose \emph{nstate} 1024 as an appropriate RNN size. We compress the RNNs by inserting layers with 512 units above and below the hidden state and then train end-to-end. Additionally, we tie the input-state and state-output weights as their transpose and quantise all the weights. At 10 MB the RNN achieves a 17\% relative hit-rate gain over the Katz-5 \emph{n}-gram, proving the utility of RNNs in memory constrained settings.

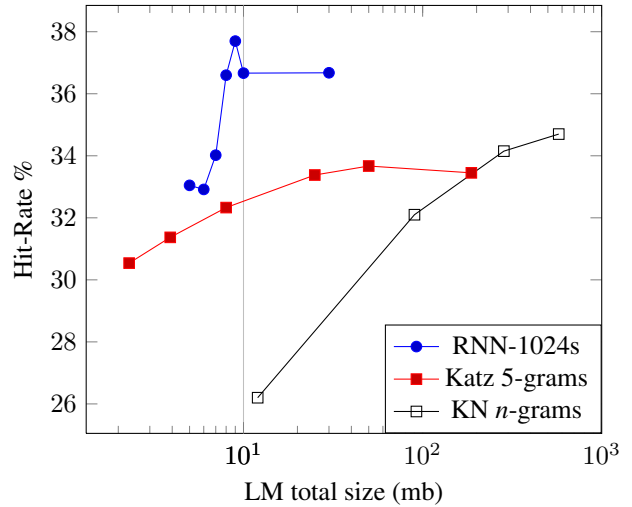
\begin{figure}
\centering
\begin{tikzpicture}
\begin{axis}[
xlabel = LM total size (mb),
ylabel = Hit-Rate \%,
xmode=log,
extra x ticks={10},
extra tick style={grid=major},
legend style={at={(1,0)},anchor=south east}
]

\addplot coordinates {
( 30, 36.6759 )
( 10, 36.6649 )
( 9, 37.6978 )
( 8, 36.5988 )
( 7, 34.0197 )
( 6, 32.9154 )
( 5, 33.0469 )
};
\addlegendentry{RNN-1024s};

\addplot coordinates {
( 187, 33.45 )
( 50, 33.67 )
( 25, 33.38 )
( 8, 32.33 )
( 3.9, 31.37 )
( 2.3, 30.54 )
};
\addlegendentry{Katz 5-grams};

\addplot[mark=square] coordinates {
( 12, 26.20 )
( 90, 32.10 )
( 284, 34.15 )
( 575, 34.70 )
};
\addlegendentry{KN \emph{n}-grams};

\end{axis}
\end{tikzpicture}
\vspace{-0.25cm}
\caption{Hit rates of Katz \emph{n}-grams with different prune thresholds, RNNs quantised to different numbers of bits per weight and KN \emph{n}-grams of increasing order. 7-bit quantisation produces an anomalously high hit-rate - we are unsure what is interacting to product this effect.}
\label{swiftkey}
\vspace{-0.25cm}
\end{figure}
\vspace{-0.25cm}
\section{Conclusion}
\vspace{-0.15cm}
\label{sec:conclusion}
We have shown that large RNNLMs can be trained efficiently on GPUs by exploiting data parallelism and minimising the number of extra parameters required during training. Such RNNs reduce the perplexity on standard benchmarks by over 40\% against 5-grams, whilst using a fraction of the parameters. We believe RNNs now offer a lower perplexity than 5-grams for any amount of training data. In addition, we showed that state of the art ASR systems can be trained with Kaldi and high-end GPUs by rescoring with an RNNLM. Despite being both compute and GPU memory bound, RNNLMs are comfortably ahead of \emph{n}-grams at present. We believe that future developments in compute power and memory capacity will further favour them.

\vfill\pagebreak

\bibliographystyle{IEEEbib}
\bibliography{bibliography}

\end{document}